\documentclass[runningheads]{llncs}
\usepackage{microtype}
\usepackage{multirow}
\usepackage[numbers]{natbib}
\usepackage{amsmath}
\usepackage{placeins}
\usepackage{amssymb}
\usepackage{pifont}
\usepackage{adjustbox}
\usepackage{array, makecell}
\usepackage{hyperref}
\usepackage[misc]{ifsym}
\AtBeginDocument{%
  }
\usepackage{multirow}
\usepackage{booktabs}     
\usepackage[table]{xcolor}

\newcommand{\best}[1]{\cellcolor{green!18}\textbf{#1}}
\newcommand{\second}[1]{\cellcolor{orange!18}\textbf{#1}}
\usepackage{xcolor}       
\usepackage{caption}
\begin{document}

\title{GeoChemAD: Benchmarking Unsupervised Geochemical Anomaly Detection for Mineral Exploration}

\titlerunning{Underwater Basket Weaving Under Extreme Pressure}

\author{Yihao Ding\inst{1} \and
Yiran Zhang\inst{1} \and
Chris Gonzalez
\inst{1} \and
Eun-Jung Holden\inst{1} \and
Wei Liu\inst{1} \thanks{Corresponding Author}}

\authorrunning{Y. Ding et al.}

\institute{The University of Western Australia, Perth, WA, Australia
\email{\{yihao.ding,yiran.zhang,chris.gonzalez,wei.liu\}@uwa.edu.au} eunjung.holden@unimelb.edu.au}

\maketitle              

\begin{abstract}
Geochemical anomaly detection plays a critical role in mineral exploration as deviations from regional geochemical baselines may indicate mineralization. Existing studies suffer from two key limitations: (1) single region scenarios which limit model generalizability; (2) proprietary datasets, which makes result reproduction unattainable. In this work, we introduce \textbf{GeoChemAD}, an open-source benchmark dataset compiled from government-led geological surveys, covering multiple regions, sampling sources, and target elements. The dataset comprises eight subsets representing diverse spatial scales and sampling conditions. To establish strong baselines, we reproduce and benchmark a range of unsupervised anomaly detection methods, including statistical models, generative and transformer-based approaches. Furthermore, we propose \textbf{GeoChemFormer}, a transformer-based framework that leverages self-supervised pretraining to learn target-element-aware geochemical representations for spatial samples. Extensive experiments demonstrate that GeoChemFormer consistently achieves superior and robust performance across all eight subsets, outperforming existing unsupervised methods in both anomaly detection accuracy and generalization capability. The proposed dataset and framework provide a foundation for reproducible research and future development in this direction. The dataset and code are publicly available at \url{https://github.com/yihaoding/geochemad}.

\end{abstract}
\section{Introduction}
Geochemical anomalies refer to baseline compositional deviations assessed at varying the local to regional spatial scales. Anomalous elemental concentrations may exhibit orders of magnitude differences in enrichment or depletion; showing changes as subtle as altering isotopic ratios or up to several weight percent in concentration \cite{carranza2008geochemical}. These anomalies are results from complex geological processes and serve as key indicators of a potential nearby mineralizing system \cite{yang2025anomaly}. 
\textbf{Geochemical Anomaly Detection (GAD)} therefore plays a critical role in mineral exploration by identifying prospective targets in both greenfields and brownfields settings, and supporting deposit targeting and environmental monitoring. In practice, geochemical anomalies are identified by analysing data derived from surface or near-surface samples to detect deviations from an expected regional baseline. However, surficial geochemical distributions represent products of primary emplacement and secondary dispersional (e.g., weathering and erosion) processes, and acquiring geochemical survey data may reflect multi-stage and multi-source metallogenic processes. These factors lead to high spatial discontinuity, uncertainty, and randomness, making it difficult to effectively leverage geological survey data for GAD.

Traditional statistical methods were initially employed by geologists to perform univariate \cite{reimann2005sub} and multivariate correlation analyses \cite{harris2001primer} among geochemical variables to identify potential mineralization. With the advent of deep learning, recent studies adopted various architectures, such as autoencoders (AE), variational autoencoders (VAE) and transformers \cite{xiong2016recognition, yu2024identification}, to construct more comprehensive and spatially-aware geochemical representations, leading to improved performance and generalizability. Based on their training paradigms, deep learning-based frameworks can be broadly categorized into supervised and unsupervised approaches. Supervised methods \cite{yang2025anomaly} rely on a sufficient number of labelled positive samples (i.e., known mineralization sites) and often struggle to generalize to regions with sparse or unknown mineralization. In contrast, unsupervised frameworks \cite{xiong2016recognition,xu2024geologically} are increasingly favoured as they can leverage large volumes of \textit{normal} samples, using reconstruction errors to detect anomalies, thereby enabling better generalization to unexplored areas.

However, several challenges remain in this direction. First, most existing GAD studies evaluate their models on privately collected datasets, which limits reproducibility and hinders fair comparison across methods. Second, majority of the research predominantly focuses on one specific region using single source geochemical data, typically sediment samples \cite{xu2024interpretable}, restricting the exploration of model generalizability across diverse scenarios, such as varying study area sizes, sampling densities, and target element types. Finally, although unsupervised learning approaches offer enhanced generalization and have gained popularity, a critical issue persists: the anomalies detected may not correspond to actual mineralization or may be unrelated to the targeted mineralize elements \cite{de2013defining}.

A primary objective of this study is to release an open-source benchmark dataset for GAD which is compiled from reliable geological surveys and encompasses multiple study regions with diverse geochemical sampling sources. This dataset could facilitate fair comparisons and rigorous evaluation of model robustness and generalizability across various application scenarios, including different spatial extents, sampling source and sampling densities. To further support the research community, we reproduce and release several widely used unsupervised anomaly detection models to serve as baselines for comparison and future exploration. Moreover, we propose a novel unsupervised GAD framework based on a transformer architecture, named GeoChemFormer. Leveraging self-supervised pretraining, GeoChemFormer learns target-element-aware geochemical representations specific to input point. The resulting latent representations can be utilized for both target element-aware anomaly detection and multivariate interpolation, addressing key limitations of existing unsupervised GAD approaches.




The contribution of this paper can be summarized into: i) We introduce GeoChemAD, a comprehensive benchmark dataset for unsupervised geochemical anomaly detection, addressing the lack of standardized benchmarking resources in this field. GeoChemAD covers multiple regions, multiple target elements, diverse sampling sources, and varying spatial scales, enabling thorough method evaluation and supporting assessment of model generalizability.
ii) We systematically reproduce and benchmark existing unsupervised GAD methods—including statistical machine learning approaches, AEs, VAEs, and Transformer-based models—on the proposed dataset, establishing the first unified performance comparison for this task.
iii) We further introduce a novel Transformer-based framework specifically designed to leverage unsupervised learning for capturing multi-element geochemical relationships and spatial correlations, yielding improved latent representations for GAD.
iv) We conduct comprehensive evaluations using various metrics and experimental settings, examining performance across thereby providing an in-depth analysis of the strengths and limitations of different unsupervised methods under diverse real-world scenarios.
\section{Related Work}

 \noindent\textbf{General Anomaly Detection.} 
 Anomaly detection refers to the task of identifying patterns in data that deviate from expected or normal behaviour, and it has attracted increasing attention across a wide range of practical applications, including industrial quality control \cite{roth2022towards} and medical diagnosis \cite{abuzaid2020identifying}. Existing anomaly detection methods can generally be categorized into supervised and unsupervised frameworks. While supervised methods \cite{ma2016supervised} often achieve strong performance, they depend heavily on labelled data and thus face challenges such as annotation cost, data imbalance, and limited generalization across regions or domains. Consequently, \textbf{unsupervised methods} have attracted increasing attention for their ability to mitigate data scarcity and generalization issues \cite{chandola2009anomaly}, ranging from traditional machine learning approaches \cite{zimek2012survey} to deep learning frameworks \cite{zong2018deep}.
 Unlike typical anomaly detection scenarios, geochemical data involve varying spatial scales, strong spatial dependence, and compositional concentration measurements shaped by complex, non-linear geological processes.


\noindent\textbf{GAD Models. \footnote{See Appendix~A for a comparison of existing GAD models.}} Early studies mainly used multivariate statistical methods, such as PCA and factor analysis \cite{jimenez1993identification}, which are limited in capturing complex non-linear patterns \cite{chen2017application}. Machine learning methods (e.g., one-class SVMs, and isolation forests) improve non-linear modelling but still struggle with the high-dimensional and spatially dependent nature of geochemical data \cite{saremi2025unsupervised}.
Recent studies aim to extract spatially aware geochemical representations using deep learning frameworks. Autoencoders (AE) and their variants \cite{xiong2016recognition} have demonstrated the ability to model compositional relationships, yet neglect spatial dependencies. Convolutional-based architectures, such as CNN \cite{zuo2024physically} and convolutional auto encoder \cite{shi2025geological,luo2025causal} preserve local spatial structure but are constrained by fixed receptive fields \cite{yu2024identification}. Graph-based models \cite{guan2022recognizing}
address long-range dependencies but are hindered by limited depth and representation capacity. Although transformers have demonstrated strong performance across many domains, their application to geochemical anomaly detection remains under-explored. Recent works have begun to leverage transformers to capture long-range spatial dependencies \cite{yu2024identification}; however, these approaches lack systematic investigation into self-supervised pretraining, which is crucial for learning generalizable geochemical patterns. 
Existing unsupervised frameworks struggle to distinguish whether detected anomalies are related to target elements or unrelated geochemical variations. Designing models that enable target-element-aware feature learning remains an open challenge. 

 \noindent\textbf{GAD Datasets.}
Most deep learning works focus on evaluating proposed frameworks within a single research area, predominantly drawing on data from the China Geological Survey and targeting gold anomalies, as illustrated by Table~\ref{tab:existing_dataset}. This narrow focus limits the generalizability of the methods, as they are not sufficiently tested across diverse scenarios such as varying sample sizes, area extents, sampling densities, or element types. Furthermore, the lack of publicly available datasets in these studies hinders fair comparison and reproducibility. Some papers also omit key metadata, further reducing the transparency and utility of their findings. 
\begin{table}[t]
\centering
\scriptsize
\caption{Summary of data sources and characteristics across different studies.}
\begin{adjustbox}{max width=\linewidth}

\begin{tabular}{l l c c c c c c}
\toprule
\textbf{Author} & \textbf{Loc.} & \textbf{\# Spl.} & \textbf{T.E.} & \textbf{Size (km$^2$)} & \textbf{\# Pos} & \textbf{Ave. Dist.} & \textbf{\# Elem.} \\
\midrule 
Luo et al. \cite{luo2025causal}         & China   & 3060   & Au  & 108000   & N/A     & 2 km     & 39 \\
Yang et al. \cite{yang2025anomaly}      & China   & 6608   & Au  & 26500    & 30*     & 2 km     & 29 \\
Shi et al. \cite{shi2025geological}     & China  & N/A    & Au  & 26500    & N/A     & 2 km     & 39 \\
Yu et al. \cite{yu2024identification}   & China   & 1585   & Au  & N/A      & 27      & 2 km     & 39 \\
Scheidt et al. \cite{scheidt2025masked} & Canada  & 9230   & Li  & 21710    & $\sim$30& 1.5 km   & 52/64 \\
Zuo et al. \cite{zuo2024physically}     & China              & N/A    & Au  & N/A      & 62      & N/A      & 39 \\
Yang et al. \cite{yang2024geologically} & China              & N/A    & Au  & N/A      & 688     & 2 km     & 39 \\
Xu et al. \cite{xu2024geologically}     & China              & N/A    & Fe  & N/A      & N/A     & 2 km     & 39 \\
Xu et al. \cite{xu2024interpretable}    & China              & N/A    & Fe  & 26500    & 19      & 2 km     & 39 \\
Xu et al. \cite{xu2024geochemical}      & China              & 2956   & Au  & 11135    & 69*     & 2 km     & 39 \\
Lin et al. \cite{lin2025domain}         & China              & N/A    & Au  & N/A      & 69*     & 2 km     & 39 \\
He et al. \cite{he2024mineral}          & China              & 5017   & Au  & N/A      & 30      & 21.5 m   & 10 \\
\bottomrule
\end{tabular}
\end{adjustbox}
\label{tab:existing_dataset}
\end{table}

\section{GeoChemAD Dataset}
\label{sec:geochemad_dataset}
\begin{figure}[t]
  \centering
  \includegraphics[width=0.85\linewidth]{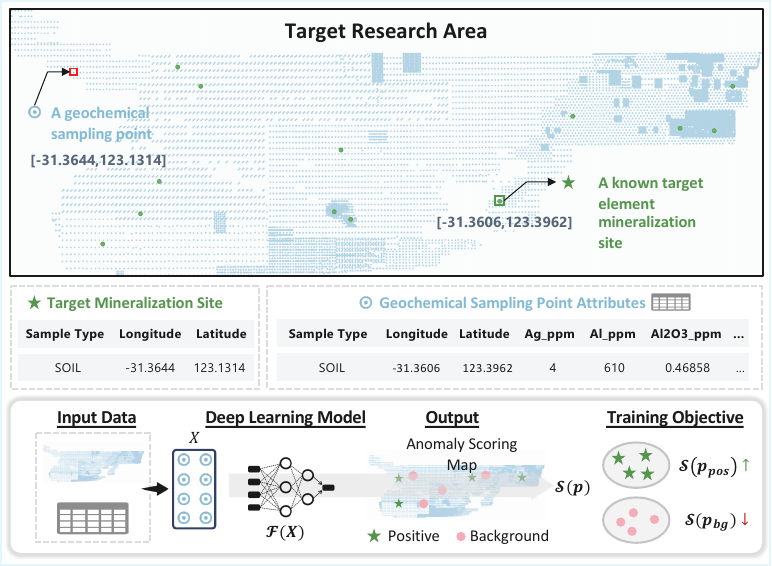}
  \caption{Geochemical Anomaly Detection Task Definition.}
  \label{fig:kdd_task_definition}
\end{figure}
\subsection{Dataset Curation}

\noindent\textbf{Data Source.} We sourced our dataset from the Geological Survey of Western Australia's (GSWA) DMPE Data in the Software Center \footnote{\url{https://dasc.dmirs.wa.gov.au/}}, an open and publicly accessible platform providing authoritative state-wide spatial data. 
Our study uses datasets from the Accelerated Geoscience Program, which include GSWA-verified geological, geochemical, and geophysical records derived from the MINEDEX database, ensuring high data quality. We focus on the CM02 Near-Surface Geochemistry subset for regional geochemical assays and the CM01 Mineralization Sites dataset for verified deposit locations. All datasets are publicly available and processed in the GDA2020 coordinate system to ensure spatial consistency.

\noindent\textbf{Data Preprocessing.} \textit{Shapefile Selection}: GeoChemAD includes geochemical sampling media from soil, sediment, and rock chip, capturing diverse surficial geochemical patterns. Each subset focuses on a specific sampling source and target element. Beyond \texttt{gold (Au)}, the primary focus of prior studies, this work also examines \texttt{tungsten (W)} and \texttt{copper (Cu)}. This reduces the limitations of single-element or single-media studies and supports a broader evaluation of the proposed framework across varied geological settings. \textit{Region Choices}: Multiple regions of interest are manually selected, whereas previous studies typically focused on a single area without considering spatial generalization. The chosen regions contain relatively evenly distributed sampling points, ensuring sufficient data density for modelling geochemical distributions with deep learning frameworks.

\noindent\textbf{Data Formatting.} 
For each subset of GeoChemAD, we provide two CSV files about geochemical samples and known mineralization sites, respectively. The geochemical sample file serves as the input for the developed unsupervised learning models and includes metadata, such as \texttt{SAMPLEID}, \texttt{SAMPLETYPE}, and spatial coordinates (\texttt{x: longitude}, \texttt{y: latitude}) in the GDA2020 datum. It also contains GSWA-postprocessed and recalculated geochemical concentrations (e.g., \texttt{Ag\_ppm}, \texttt{Au\_ppb} and etc.). The exact number of chemical elements vary by subset and recorded in Table~\ref{tab:geochem_datasets}.
Notably, the data may include abnormal values (e.g., \texttt{-9999}, \texttt{-0.5}), which are retained to preserve data integrity requiring appropriate preprocessing techniques to handle.
Known mineralization site CSV files contain metadata fields such as \texttt{SiteID} and \texttt{ProjectID}, along with spatial coordinates \texttt{(x, y)}. This file is intended as a reference dataset to evaluate the model's ability to detect or predict mineralization zones.

\subsection{Task Definition}
Given a set of geochemical samples $\mathbb{S}=\{S_i\}$, where each sample is represented as 
$S_i=\{p_i, \mathbf{u}_i\}$ with position $p_i=(\text{x}_i, \text{y}_i)$ denoting its spatial 
location and $\mathbf{u}_i=(u_{i,1}, u_{i,2}, \ldots, u_{i,c})$ denoting its multi-element 
geochemical concentrations, and a set of known deposits $D=\{d_j\}$ associated 
with a specific target element, the objective is to 
train a machine learning framework $\mathcal{F}$ in an unsupervised manner and use the trained 
model to produce an anomaly-scoring function $\mathcal{V}$. 
We expect the anomaly function to assign higher scores 
$\mathcal{V}(p_{\text{pos}})$ to the locations of known target-element mineralization 
sites, and lower scores $\mathcal{V}(p_{\text{bg}})$ to background samples. 
The resulting continuous anomaly score map $\mathcal{V}(p)$ provides a quantitative 
representation of mineralization potential across the study area, enabling the 
identification of new prospective mineralized zones. Following previous studies, background samples are randomly selected from the research area. \textbf{Evaluation Metrics:} Following previous studies, model performance is primarily evaluated using the Area Under the ROC Curve (AUC), averaged over 20 runs with repeated background sampling to ensure robustness; additional metrics assessing spatial targeting and spatial consistency are provided in the Appendix B.

\subsection{Dataset Analysis}

\begin{figure*}[h]
  \centering
  \includegraphics[width=\linewidth]{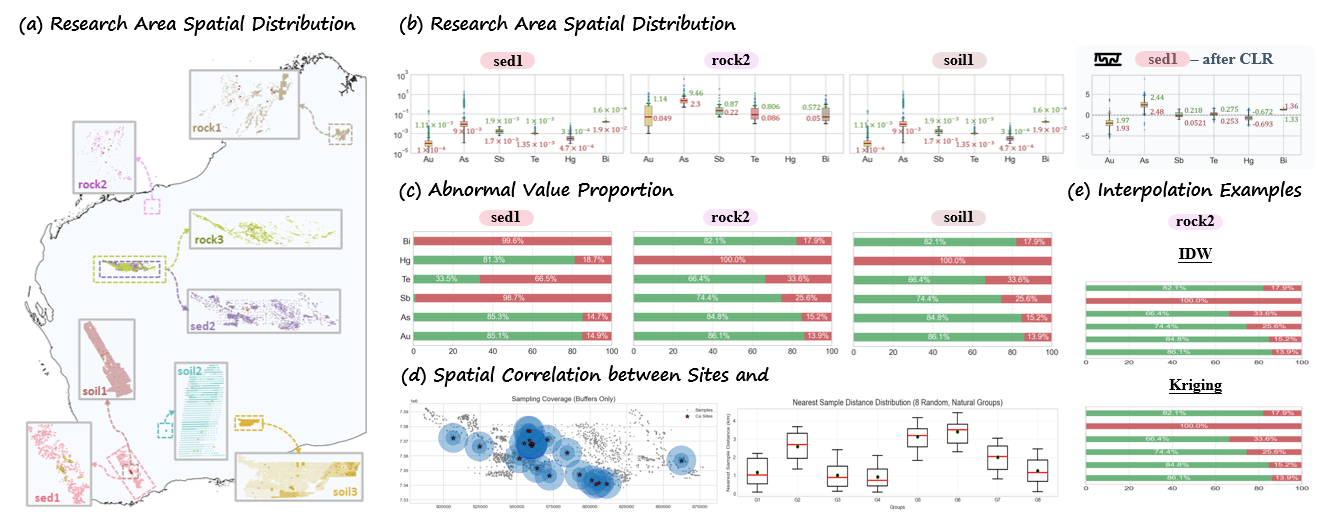}
  \caption{Dataset analysis of the GeoChemAD dataset, including (a) the distribution of research areas, (b–c) the distribution of element concentration values, (d) the spatial correlation between mineralization sites and geochemical samples, and (e) examples of spatial interpolation.}
  \label{fig:dataset_analysis}
\end{figure*}

\textbf{Basic Statistics.}
We present the basic statistics of the eight GeoChemAD subsets in Table~\ref{tab:geochem_datasets}, including sampling source, spatial coverage, sampling density, and target element details. Unlike prior studies that often rely on a single sampling medium, typically sediment, GeoChemAD incorporates data from diverse surface geochemical surveys: two sediment, three rock chip, and three soil subsets. This variety allows for robust assessment of model generalizability across different geological settings.
Beyond sampling source diversity, GeoChemAD spans a wide range of spatial scales, with sampling point counts and area sizes varying from approximately 6~km$^2$ to 8,500~km$^2$, supporting evaluations under both sparse and dense sampling conditions. Additionally, while many existing studies focus narrowly on precious metals such as gold, GeoChemAD expands the scope to include targeting of multiple commodities, \texttt{Au}, \texttt{Cu}, \texttt{Ni}, and \texttt{W}, across different regions and deposit styles in Western Australia (see Fig.~\ref{fig:dataset_analysis} (a). Each subset includes a sufficient number of known deposits, ensuring meaningful and consistent evaluation of anomaly detection performance.

\begin{table}[ht]
\centering
\caption{Summary of Multi-source Geochemical Datasets and Anomaly Points}
\setlength{\tabcolsep}{1pt} 
\renewcommand{\arraystretch}{1} 
\begin{adjustbox}{max width=0.88\linewidth}

\begin{tabular}{lccccccc}
\toprule
\textbf{Source} & \textbf{ID} & \textbf{\#Samp} & \textbf{Area (km$^2$)} & \textbf{Dist (km)} & \textbf{Elem} & \textbf{AnomPts} & \textbf{\# Elem} \\
\midrule
\multirow{2}{*}{Sediment} 
 & sed1 & 1392 & $\sim$8523 & 2.48 & \texttt{Au} & 32 & 124\\
 & sed2 & 2994 & $\sim$6671 & 1.49 & Cu & 21 & 124\\
\midrule
\multirow{3}{*}{RockChip} 
 & rock1 & 3790 & $\sim$3177 & 0.91 & W & 7 & 124\\
 & rock2 & 224 & $\sim$2 & 0.09 & Au & 12 & 124\\
 & rock3 & 9624 & $\sim$7423 & 0.88 & Cu & 21 & 124\\
\midrule
\multirow{3}{*}{Soil} 
 & soil1 & 2734 & $\sim$5.7 & 0.05 & Au & 14 & 126\\
 & soil2 & 5163 & $\sim$57 & 0.11 & Au & 17 & 126\\
 & soil3 & 21040 & $\sim$2018 & 0.31 & Ni & 13 & 126\\
\bottomrule
\end{tabular}
\end{adjustbox}
\label{tab:geochem_datasets}
\end{table}

\noindent\textbf{Geochemical Concentration Distribution.}
To better understand the geochemical characteristics of GeoChemAD, we analyze element concentration patterns in Fig.~\ref{fig:dataset_analysis} (b) and (c).
As shown in Fig.~\ref{fig:dataset_analysis} (b), element concentrations vary substantially across sample media, with the same element exhibiting apparently different value ranges. 
Rock samples generally show higher and more compact concentration levels, sediment samples span wider ranges with pronounced high-value tails, while soil samples mainly display lower and more constrained backgrounds. This diversity in concentration scales and distributional forms motivates the inclusion of multiple sample media, enabling evaluation under heterogeneous geochemical conditions. To further illustrate distributional differences, we report concentration distributions of the same elements within the ``\textit{sed1}'' subset, showing that applying CLR transformation effectively mitigates extreme values, resulting in more comparable and normalized distributions.

\noindent\textbf{Spatial Correlation between samples and sites.}
In Fig.~\ref{fig:dataset_analysis} (d), the left panel visualizes the spatial relationship between deposit sites and surrounding samples using multiple concentric buffer zones, showing that each target \texttt{Cu} site (``\textit{sed2}'') is adequately covered by samples at different spatial scales. The right panel reports the distribution of sample counts within a buffer defined as five times the average sampling distance, indicating consistent and representative sampling coverage across different subsets and sample media.

\section{GeoChemFormer}
\begin{figure*}[t]
  \centering
  \includegraphics[width=0.9\linewidth]{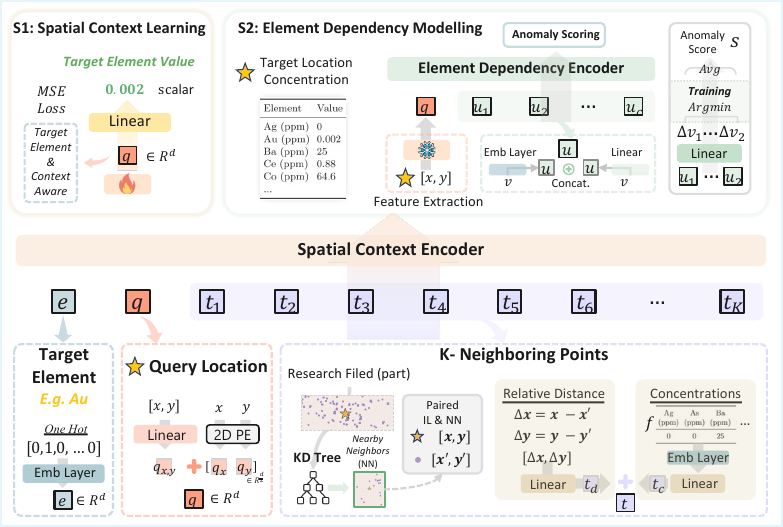}
  \caption{GeoChemFormer model architecture, first learns spatially informed representations from neighbourhood context, and then detects anomalies by modelling dependencies among elemental concentrations conditioned on this spatial context.}
  \label{fig:geochemformer}
\end{figure*}
The challenges of geochemical data understanding result from high-dimensional multi-element concentrations, strong compositional dependencies, and spatially structured geological processes. Consequently, effective mineral exploration requires models that capture both \textit{spatial geochemical context} and \textit{element dependency patterns}. To address this, we propose \textbf{GeoChemFormer}, a pretrained transformer framework for GAD, as shown in Fig.~\ref{fig:geochemformer}.

\subsection{Spatial Context Learning}

The first stage aims to learn a latent \textbf{Spatial Context} (SC) representation that captures local geochemical co-variation patterns around an query location. 
Rather than using the geochemical value at the query location, the model predicts the target element concentration solely from its surrounding neighbours, forcing the model to learn geological context from nearby samples.

\noindent\textbf{Neighbourhood Construction.}
For each query location $p_i=(x_i,y_i)$, we retrieve its $K$ nearest neighbouring samples using a KD-tree. 
Each neighbour $p_j$ contributes both spatial and geochemical information. 
Specifically, the neighbour token is defined as: $
\mathbf{t}_j =
\left[
\Delta x_j,\,
\Delta y_j,\,
\mathbf{f}_j
\right]$, where $(\Delta x_j,\Delta y_j)=(x_j-x_i,y_j-y_i)$ denotes the relative spatial offset and $\mathbf{f}_j \in \mathbb{R}^{c}$ represents the preprocessed concentration vector of $c$ geochemical element types.

\noindent\textbf{Spatial Context Encoder.}
GeoChemFormer processes a sequence of tokens consisting of
$\mathcal{S} =
[\mathbf{e},\mathbf{q}_i,\mathbf{t}_1,\ldots,\mathbf{t}_K]$, where $\mathbf{e}$ denotes the target-element token representing the interest commodity, $\mathbf{q}_i$ encodes the spatial coordinates of the query location, and $\mathbf{t}_j$ represents the $K$ nearest neighbouring sample feature. $K$ is a hyperparameter controlling the number of neighbours. The target token $\mathbf{q}_i$ is obtained by linearly projecting $(x_i,y_i)$ and adding a 2D positional encoding. 
Neighbour tokens are formed by projecting the neighbour feature vector and incorporating relative spatial offsets.
Then, the token sequence is processed by an $L$-layer Transformer encoder $\mathcal{E}_{\text{scl}}$ encodes the spatial context sequence $\mathcal{S}$, and the spatially informed query location representation is obtained from the second token:
$\mathbf{q}_i' = \mathcal{E}_{\text{scl}}(\mathcal{S})[1].$ This latent vector $\mathbf{q}_i'$ summarises the geochemical context surrounding location $p_i$.

\noindent\textbf{Spatial Context Learning Objective.}
During spatial context learning, the model predicts the concentration of the target element at the query location using neighbouring samples:
$\hat{y}_i = W_{\text{pred}}\mathbf{q}_i'$. $W_{\text{pred}}$ and the following $W$ denote linear projections. 
The model is trained by minimising the mean squared error:

\begin{equation}
\mathcal{L}_{\text{sc}} =
\frac{1}{N}\sum_{i=1}^{N}(\hat{y}_i-y_i)^2
\end{equation}

After spatial context learning, the latent representation $\mathbf{q}_i'$ is extracted for every sample and used as geo-context information in the anomaly detection stage.

\subsection{Element Dependency Modelling for Anomaly Detection}

The second stage detects anomalous geochemical signatures by modelling dependencies among elemental concentrations conditioned on the learned spatial context. For each sample, the input sequence consists of a \textbf{geo-context token} derived from the spatial representation $\mathbf{q}_i'$, followed by a sequence of element tokens: $\mathcal{S}' = [W_g\mathbf{q}_i',\mathbf{u}_1,\ldots,\mathbf{u}_c]$.
Each element token $\mathbf{u}_c$ represents a specific geochemical element and is constructed by combining an element identity embedding with the corresponding scalar concentration value:
$
\mathbf{u}_c =
W_e
\left[
\mathrm{Embed}(c)
\;\middle|\;
W_v x_{i,c}
\right]
$
This design enables the model to explicitly capture inter-element relationships.

\noindent\textbf{Element Dependency Encoder}
The token sequence is processed by a Transformer encoder that learns dependencies among geochemical elements conditioned on the spatial context token. 
A shared linear decoder then reconstructs the concentration of each element:
$
\hat{\mathbf{x}}_i =
W_d \cdot
\mathrm{TransformerEncoder}(\mathcal{S}')
[1:c+1]
$.

\noindent\textbf{Anomaly Scoring}
The anomaly score is defined as the mean squared reconstruction error across all elements:

\begin{equation}
s_i =
\frac{1}{C}
\sum_{c=1}^{C}
(x_{i,c}-\hat{x}_{i,c})^2
\end{equation}

Samples whose geochemical signatures deviate from the learned element dependency patterns produce larger reconstruction errors and are therefore assigned higher anomaly scores.

\section{Evaluation Setup}
\label{sec:evaluation_setup}
\textbf{Preprocessing of Geochemical Data for Model Input.} 
To facilitate geochemical concentration modelling, several preprocessing procedures are applied depending on the requirements of the selected model.
\textbf{(a)}\textit{Abnormal Values.} 
Geochemical surveys frequently contain abnormal entries, including missing, negative, or zero values. These are typically handled by removing the affected samples or replacing them with half of the detection limit to maintain the statistical structure of the dataset.
\textbf{(b)}\textit{Closure Issue}.
Because geochemical concentrations are compositional and constrained by closure, spurious correlations may arise. Log-ratio transformations are therefore commonly applied, like \textit{Centered Log-Ratio (CLR)} and \textit{Isometric Log-Ratio (ILR)} transformations.
\textbf{(c)}\textit{Feature Selection.} 
Geochemical surveys are multivariate and contain concentrations of numerous elements. Feature selection can significantly influence GAD depending on the target element, area of interest, and source variations. In this study, several feature selection approaches are compared, including manual selection, PCA, causal discovery methods, and LLM-assisted methods.
\textbf{(d)}\textit{Interpolation.} 
When raster-based model inputs are required, spatial interpolation methods such as \textit{Inverse Distance Weighting (IDW)} and \textit{Kriging} are used to convert discrete geochemical samples into raster images. Detailed descriptions with an illustration diagram of these preprocessing procedures are provided in the Appendix C.

\noindent\textbf{Baseline Models.}
To provide a comprehensive evaluation, we include a diverse set of unsupervised anomaly detection baselines spanning statistical methods, classical machine learning models, deep generative frameworks, and transformer-based architectures (Table~3). Statistical approaches include Z-score and Mahalanobis Distance, which detect anomalies based on deviations from global data distributions, and kNN, which measures local neighborhood sparsity. Classical machine learning baselines include Isolation Forest and One-Class SVM, which identify anomalies through isolation-based partitioning and boundary learning, respectively. For deep learning baselines, we adopt reconstruction-based generative models, including AutoEncoder (AE), Variational AutoEncoder (VAE), and their extensions such as VAE-GAN, cascade-GAN variants, and diffusion-based models. These approaches detect anomalies through reconstruction or generation errors in the learned latent space. Finally, we include two transformer-based models: a vanilla transformer baseline (T1) and our proposed \textbf{GeoChemFormer} (T2), which leverage attention mechanisms to model spatial relationships and contextual dependencies among geochemical samples.
All baseline models are implemented in PyTorch on a single
NVIDIA RTX A6000 (48\,GB) GPU; full implementation details and hyperparameters are listed in
Appendix~D.

\section{Results and Discussion}
\label{sec:results}
\begin{table*}[t]
\centering
\footnotesize
\setlength{\tabcolsep}{4pt}
\caption{
Performance comparison of anomaly detection methods across geochemical datasets. 
Abbreviations: 
ZS = Z-score; 
MD = Mahalanobis Distance; 
KNN = k-Nearest Neighbors; 
IF = Isolation Forest; 
OSVM = One-Class SVM; 
AE = AutoEncoder; 
VAE = Variational AutoEncoder; 
VAE-G = VAE-GAN; 
VAE-C = VAE-cascade-GAN; 
VAE-D = VAE-Diffusion; 
T1 and T2 denote two Transformer-based models (Vanilla and Our GeoChemFormer).
}
\begin{adjustbox}{max width=\textwidth}
\begin{tabular}{l|ccc|cc|ccccccc}
\toprule
\multirow{2}{*}{Area} 
& \multicolumn{3}{c|}{Statistical} 
& \multicolumn{2}{c|}{Classical ML} 
& \multicolumn{7}{c}{Deep Generative / Transformer} \\
\cmidrule(lr){2-4}
\cmidrule(lr){5-6}
\cmidrule(lr){7-13}
& ZS 
& MD 
& KNN 
& IF 
& OSVM 
& AE 
& VAE 
& VAE-G 
& VAE-CG
& VAE-D 
& T1
& T2 \\
\midrule
sed1  & 0.3220 & 0.5224 & 0.6169 & 0.6067 & 0.5410 & 0.5851 & 0.5441 & 0.6843 & 0.7103 & 0.6948 & \second{0.7111} & \best{0.7228} \\
sed2  & 0.7101 & 0.7038 & 0.6938 & \second{0.7707} & 0.7471 & \best{0.7991} & 0.5204 & 0.7586 & 0.7002 & 0.6234 & 0.7484 & 0.7339 \\
rock1 & 0.5773 & 0.5109 & 0.5086 & 0.6836 & 0.5625 & 0.5516 & 0.6672 & 0.6953 & 0.6281 & 0.6391 & \second{0.7031} & \best{0.7844} \\
rock2 & 0.3056 & 0.4778 & 0.5562 & 0.5151 & 0.4870 & \best{0.9185} & \second{0.8704} & 0.7901 & 0.7593 & 0.6086 & 0.7444 & 0.8050 \\
rock3 & 0.5813 & 0.6761 & 0.7424 & 0.5475 & 0.6869 & \best{0.8446} & 0.6694 & 0.7527 & 0.6351 & 0.6492 & \second{0.7986} & 0.7302 \\
soil1 & 0.5209 & 0.5107 & 0.6282 & 0.5584 & 0.4944 & 0.5934 & 0.5331 & 0.7124 & 0.6074 & 0.5992 & \second{0.7242} & \best{0.8704} \\
soil2 & 0.6122 & 0.5247 & 0.5733 & 0.5207 & 0.5321 & \second{0.7900} & 0.6889 & \best{0.8138} & 0.6481 & 0.5965 & 0.6780 & 0.6896 \\
soil3 & 0.3793 & 0.3447 & 0.3527 & 0.4441 & 0.4411 & 0.5544 & 0.6337 & 0.6160 & \second{0.6509} & 0.6438 & 0.6101 & \best{0.8334} \\
\midrule
Avg.  & 0.5011 & 0.5339 & 0.5840 & 0.5809 & 0.5615 & 0.7046 & 0.6409 & \second{0.7279} & 0.6674 & 0.6318 & 0.7147 & \best{0.7712} \\
Var.  & 0.0220 & 0.0128 & 0.0143 & 0.0108 & 0.0109 & 0.0220 & 0.0131 & 0.0041 & 0.0026 & 0.0011 & 0.0031 & 0.0039 \\
\bottomrule
\end{tabular}
\end{adjustbox}
\label{tab:main_performance}
\end{table*}

\subsection{Overall Performance}

Table~\ref{tab:main_performance} presents a systematic comparison across statistical, classical machine learning, and deep generative / transformer-based models, to evaluating their effectiveness for geochemical anomaly detection. Overall, statistical approaches (ZS, MD, KNN) show limited performance, with an average AUC of 0.5--0.58, reflecting their inability to capture complex geochemical relationships. Classical machine learning methods (IF, OSVM) provide moderate improvements, indicating that non-linear modelling partially benefits GAD. 
Deep generative models generally achieve stronger results, demonstrating the advantage of representation learning for modelling high-dimensional geochemical distributions. For example, AE achieves an average AUC of 0.7046, while VAE-G further improves performance to 0.7279. Transformer-based models further enhance detection by modelling global dependencies. Notably, GeoChemFormer (T2) achieves the best overall performance with an average AUC of \textbf{0.7712}, outperforming the vanilla transformer (T1, 0.7147) and other deep generative baselines. This highlights the effectiveness of learning \textit{spatial context} and \textit{element dependency patterns}.
Nevertheless, performance variations across datasets (e.g., AE achieving 0.9185 on \textit{rock2}, while T2 achieves 0.8334 on \textit{soil3}) suggest that different models may perform better under specific geological settings. This indicates that geochemical anomaly detection remains sensitive to dataset characteristics, and further work is needed to improve robustness across diverse exploration scenarios.

\subsection{Effects of Geochemical Data Preprocessing}

\begin{figure}[h]
  \centering
  \includegraphics[width=0.795\linewidth]{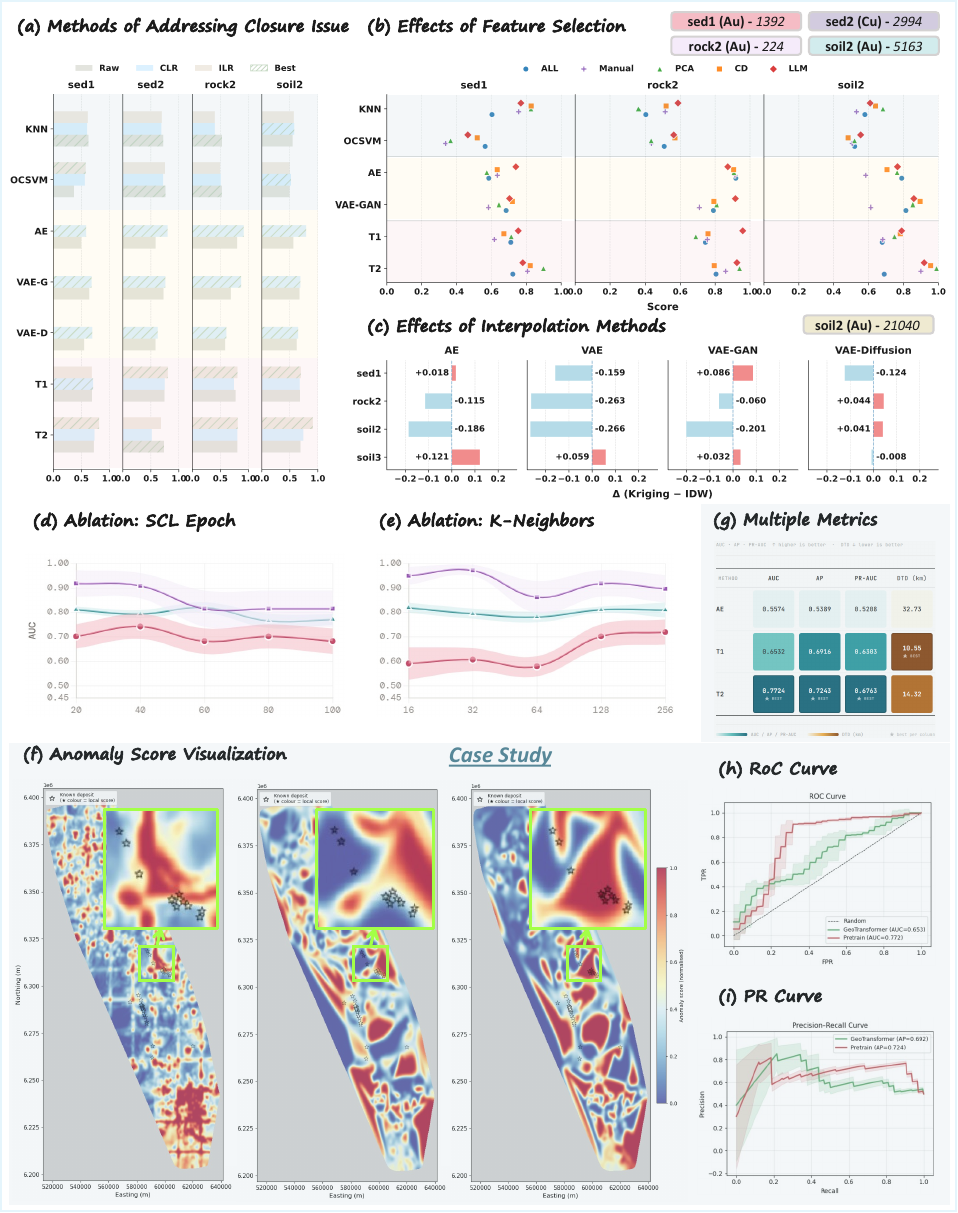}
  \caption{Additional experiments about effects of various data preprocessing strategies (a) to (c), ablation testing (d) and (e), case studies (f) to (i).}
  \label{fig:result_analysis}
\end{figure}

\noindent\textbf{Effectiveness of Approaches for Addressing the Closure Issue.}
To evaluate compositional transformations, we compare raw concentrations (Raw), CLR and ILR transformations (Fig.~\ref{fig:result_analysis} (a)). Overall, ILR achieves the best average performance (0.6788), closely followed by CLR (0.6771), while Raw inputs perform worst (0.6406), indicating that addressing the compositional closure effect generally improves anomaly detection. From the dataset perspective, log-ratio transformations provide benefits for Au-targeted datasets, improving performance by approximately +0.05 compared to raw inputs, while having little effect on the Cu-targeted dataset. From the model perspective, sensitivity to compositional transformations varies across architectures. Classical models respond differently to compositional transformations, where KNN shows slight performance decreases after transformation, whereas OCSVM benefits from both CLR and ILR. AE-based models (AE, VAE-GAN, and VAE-Diffusion) generally benefit from CLR transformations, while transformer-based models achieve the best performance with ILR consistently. These results suggest that compositional transformations are generally beneficial, but the optimal strategy depends on both the dataset and the model architecture.

\noindent\textbf{Effectiveness of Feature Selection Approaches.}
We evaluate five feature selection strategies (Section \ref{sec:evaluation_setup}) across multiple anomaly detection models, including not applying feature selection to use all available elements (NA), manual selection based on domain knowledge (Manual), PCA, causal discovery (CD), and large language model assisted selection (LLM). Experiments on three Au-targeted subsets show that automated approaches (PCA, CD, and LLM) generally achieve more stable and higher performance than manual selection, suggesting that they capture latent geochemical relationships relevant to \texttt{Au} mineralization. In particular, the LLM-based strategy achieves the best overall results, with the highest average score (0.7412) compared to manual selection (0.6419). From the model perspective, classical models such as KNN and OCSVM remain sensitive to feature selection, although their optimal strategies differ across methods, with both statistical reductions and automated selections often outperforming manually selected subsets. In contrast, deep learning models, including AE and VAE-GAN, are more robust to feature choices and benefit from automated selection strategies. Transformer-based models achieve the best overall performance, with LLM-selected features performing best for the standard Transformer, while PCA-based features yield the highest score (0.9427) for our proposed model. Overall, these results suggest that feature selection effectiveness depends on the model architecture, with classical models showing higher sensitivity to feature selection strategies, while deep models better leverage informative feature subsets.

\noindent\textbf{Effectiveness of Interpolation Approaches.}
We compare inverse distance weighting (IDW) and Kriging across datasets with different sampling distances and target elements (Fig.~\ref{fig:result_analysis} (c)). From the dataset perspective, interpolation effectiveness varies with both sampling density and target element. For Au-targeted datasets, IDW generally provides slightly better performance, whereas Kriging performs better for the Ni-targeted dataset. Sampling distance also influences the impact of interpolation: the 1 km dataset achieves the highest overall performance (average score 0.8080), while denser sampling datasets (100 m) exhibit larger performance differences between interpolation strategies, with a maximum gap of 0.2664. From the model perspective, interpolation sensitivity varies across architectures, with VAE showing the largest performance variation between IDW and Kriging, while VAE-Diffusion is the most robust. Overall, these results suggest that interpolation effectiveness is primarily driven by data characteristics rather than model architecture. More details in Appendix E.

\subsection{Ablation Testing}
\noindent \textbf{Effect of Spaital Context Learning (SCL) Epochs}. According to Fig.~\ref{fig:result_analysis}, GeoChemFormer reaches its best AUC within a relatively short pre-training budget: \textit{rock2} peaks at 20 epochs (0.919) and sed1 at 40 epochs (0.743), while \textit{soil2} benefits slightly from longer training, peaking at 60 epochs (0.821). These results demonstrate the effectiveness of SCL in capturing geochemical relationships from neighbouring samples. However, the optimal training duration depends on the complexity of spatial distributions and elemental interactions, which may require additional epochs to fully learn stable representations in diverse scenarios.

\noindent \textbf{Effect of neighbourhood size K}. The sensitivity to K varies considerably across datasets. sed1 improves monotonically as K grows from 16 to 256 (0.591 to 0.720), indicating that sediment anomalies benefit from a broader spatial context. In contrast, rock2 and soil2 peak at small K values (32 and 16, respectively) and decline thereafter, implying that tight local neighbourhoods better capture the sharp geochemical contrasts characteristic of hard-rock and soil targets. Overall, a moderate K of 128 provides a reasonable cross-dataset compromise, and all datasets exhibit higher variance at the extremes, underscoring the importance of K selection for stable anomaly detection.

\subsection{Case Studies}

Fig.~\ref{fig:result_analysis} (f) presents anomaly score visualizations for different models over the same study region. Compared with other approaches, GeoChemFormer produces anomaly patterns that align more closely with known mineralization sites. High anomaly scores are concentrated around deposit clusters rather than scattered across unrelated areas, indicating that the model effectively captures spatial geochemical context. The zoomed-in regions further demonstrate that GeoChemFormer generates smoother and more geologically coherent anomaly structures, highlighting its ability to better localize prospective mineralization zones.

In addition, the results in Fig.~\ref{fig:result_analysis} (f) show consistent improvements of GeoChemFormer across multiple evaluation metrics. In particular, it achieves the highest AUC and competitive AP and PR-AUC scores, indicating stronger discrimination between anomalous and background samples. Meanwhile, spatial metrics such as distance to deposits (DTD) also demonstrate that the predicted anomalies are geographically closer to known mineralization sites. Overall, these trends suggest that GeoChemFormer not only improves classification performance but also produces more spatially meaningful anomaly predictions for mineral exploration.
\section{Conclusion}
In this work, we introduce GeoChemAD, a comprehensive benchmark dataset for unsupervised geochemical anomaly detection that addresses the lack of publicly available datasets covering diverse sampling sources, spatial scales, and target elements. We further provide a systematic benchmark of existing statistical, machine learning, and deep learning approaches, establishing strong baselines for future research. Building on these insights, we propose GeoChemFormer, a transformer-based framework that jointly models spatial geochemical context and element dependency patterns through self-supervised learning. Extensive experiments demonstrate that GeoChemFormer achieves superior and more stable performance across diverse datasets. We hope that GeoChemAD and the proposed framework will facilitate reproducible research and inspire further advances in AI-driven mineral exploration.

\bibliographystyle{ACM-Reference-Format}
\bibliography{sample-base}

\appendix

%
%
%
%

\end{document}